\documentclass[runningheads]{llncs}

\usepackage[year=2024,ID=5143]{eccv}
\usepackage{eccv}

\usepackage{eccvabbrv}

\usepackage{graphicx}
\usepackage{booktabs}

\usepackage[accsupp]{axessibility}  

\usepackage{hyperref}

\usepackage{orcidlink}

\usepackage{pifont}   
\usepackage{xcolor}   
\usepackage{multirow}
\usepackage{graphicx}
\usepackage{amssymb}
\newcommand{\redcross}{\textcolor{red}{\ding{55}}}
\newcommand{\greentick}{{\color{green}\ding{51}}}

\usepackage{newfloat}
\usepackage{listings}
\DeclareCaptionStyle{ruled}{labelfont=normalfont,labelsep=colon,strut=off} 
\usepackage{tablefootnote}

\begin{document}

\title{Platypus: A Generalized Specialist Model for Reading Text in Various Forms} 

\titlerunning{Platypus: A Generalized Specialist Model for Reading Text in Various Forms}

\author{Peng Wang\thanks{Equal contribution. $\dagger$ Corresponding author.}\orcidlink{0009-0001-8617-1550} \and
Zhaohai Li$^{\star}$\orcidlink{0000-0002-7704-3231} \and 
Jun Tang$^{\star}$\orcidlink{0009-0006-3949-007X} \and Humen Zhong\orcidlink{0009-0002-8676-0811} \and Fei Huang\orcidlink{0000-0002-3709-5053} \and Zhibo Yang$^{\dagger}$\orcidlink{0000-0003-2343-7750}  \and Cong Yao$^{\dagger}$\orcidlink{0000-0001-6564-4796} }

\authorrunning{P. Wang et al.}

\institute{Alibaba Group, Beijing, China\\
\email{\{wdp0072012,tjbestehen,yangzhibo450,yaocong2010\}@gmail.com \\ zhaohai.li@foxmail.com,zhonghumen@smail.nju.edu.cn,f.huang@alibaba-inc.com}}

\maketitle

\begin{abstract}

Reading text from images (either natural scenes or documents) has been a long-standing research topic for decades, due to the high technical challenge and wide application range. Previously, individual specialist models are developed to tackle the sub-tasks of text reading (e.g., scene text recognition, handwritten text recognition and mathematical expression recognition). However, such specialist models usually cannot effectively generalize across different sub-tasks. Recently, generalist models (such as GPT-4V), trained on tremendous data in a unified way, have shown enormous potential in reading text in various scenarios, but with the drawbacks of limited accuracy and low efficiency. In this work, we propose Platypus, a generalized specialist model for text reading. Specifically, Platypus combines the best of both worlds: being able recognize text of various forms with a single unified architecture, while achieving excellent accuracy and high efficiency. To better exploit the advantage of Platypus, we also construct a text reading dataset (called Worms), the images of which are curated from previous datasets and partially re-labeled. Experiments on standard benchmarks demonstrate the effectiveness and superiority of the proposed Platypus model. Model and data will be made publicly available at \href{https://github.com/AlibabaResearch/AdvancedLiterateMachinery/tree/main/OCR/Platypus}{\textcolor{magenta}{AdvancedLiterateMachinery}}.


\end{abstract}

\begin{figure}[h]
    \centering
    \includegraphics[width=0.98\textwidth]{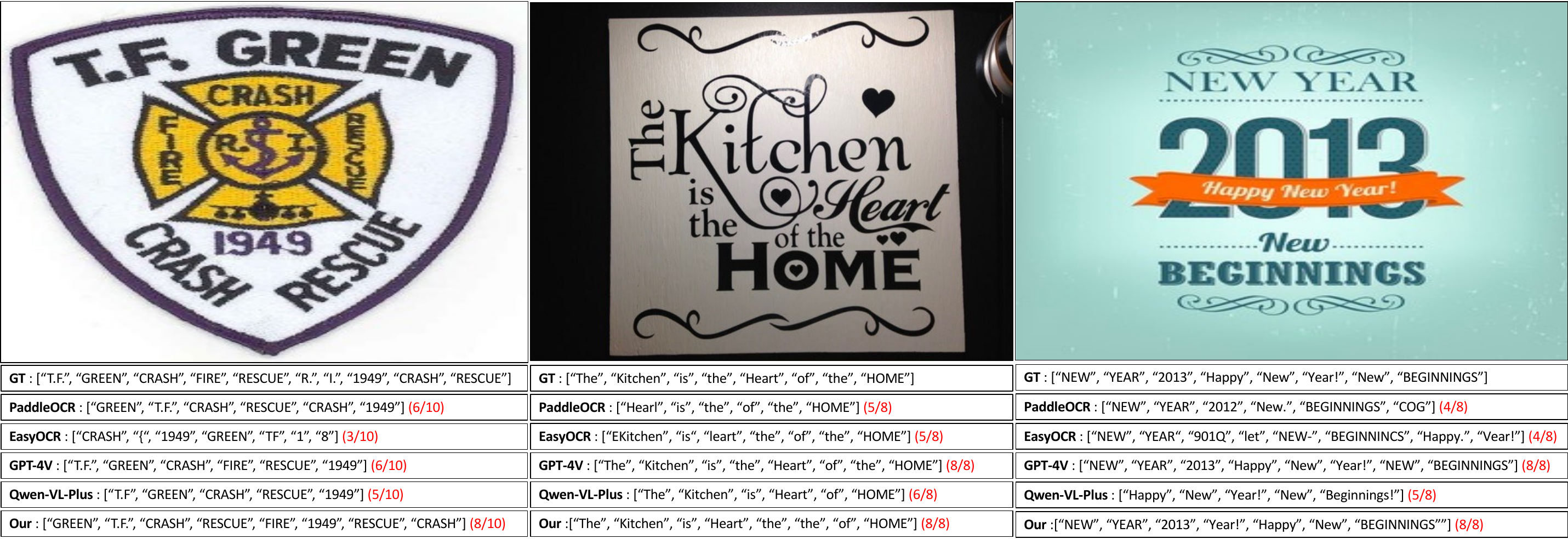}
    \caption{Comparative cases of Platypus against MLLMs (GPT-4V~\cite{2023GPT4VisionSC}, Qwen-VL-Plus~\cite{Bai2023QwenVLAV}) and OCR tools (PaddleOCR, EasyOCR) on CAT Benchmark, highlighting word accuracy ratio (red brackets) and Platypus's RAT performance.}
    \label{fig:laionocr}
\end{figure}

\section{Introduction}
\label{sec:intro}

The task of reading text from images, whether in natural scenes or documents, has evolved substantially to support a broad spectrum of applications, from archiving historical documents to real-time translation services~\cite{Zhu2015SceneTD,Long2018SceneTD}. While traditionally categorized within the domain of Optical Character Recognition (OCR), the ambition of this task extends beyond the conventional bounds of OCR to encompass a wider notion of text reading. 

The field's progression has seen the advent of models specialized for distinct text reading sub-domains, such as scene text recognition~\cite{2016pami_shi_crnn,2018pami_shi_aster,2019iccv_baek_whatiswrong}, handwritten note conversion~\cite{Yang2022DiG,Aberdam2020SeqCLR,Nuriel2021TextAdaINPA,Liu2022Persec}, and formula decoding~\cite{HME100K}. Despite their proficiency within specific tasks, these specialized models struggle with the unpredictability of text in broader scenarios~\cite{2019iccv_baek_whatiswrong,OOV}. 

Traditional text reading systems typically consist of two distinct stages: text detection~\cite{2017cvpr_zhou_east,2019cvpr_wang_psenet,2019iccv_wang_pan,2017cvpr_shi_seglink,2019pr_tang_seglink++,2018eccv_long_textsnake,2021cvpr_he_most} followed by text recognition~\cite{2016pami_shi_crnn,2018pami_shi_aster,Bautista2022PARSeq,Fang2021ABINet}. The overall performance is highly reliant on the accuracy of the initial text detection phase, creating a cascading effect when errors occur. Moreover, the emergence of Multi-modal Large Language Models(MLLMs) has contributed a new dimension to the field by offering models with broad text reading capabilities~\cite{Shi2023ExploringOCR,Bai2023QwenVLAV,Ye2023mPLUGDocOwlMM,Li2023MonkeyIR}, albeit with the trade-off of computational efficiency and specialized precision~\cite{Liu2023OnTH}.

In response to these challenges, we present Platypus, a novel generalized specialist model for text reading that offers a comprehensive solution for diverse text reading scenarios. Platypus embodies the amalgamation of high-precision specialized text reading models with the wide-ranging adaptability of multimodal approaches. The name Platypus is inspired by the animal known for its unique combination of traits; similarly, our Platypus model harmonizes the specificity of specialized models with the versatility of generalist approaches, making it adept at tackling the heterogeneity of text reading tasks.

Fig.~\ref{fig:laionocr} shows Platypus's qualitative comparisons with MLLMs and open-source OCR pipelines on the Curated Artistic Text (CAT) Benchmark. It highlights Platypus's superior accuracy in identifying words from artistically rendered text images, a challenging scenario for many contemporary OCR systems.

\begin{figure}[t!]
    \centering
    \includegraphics[width=0.95\textwidth]{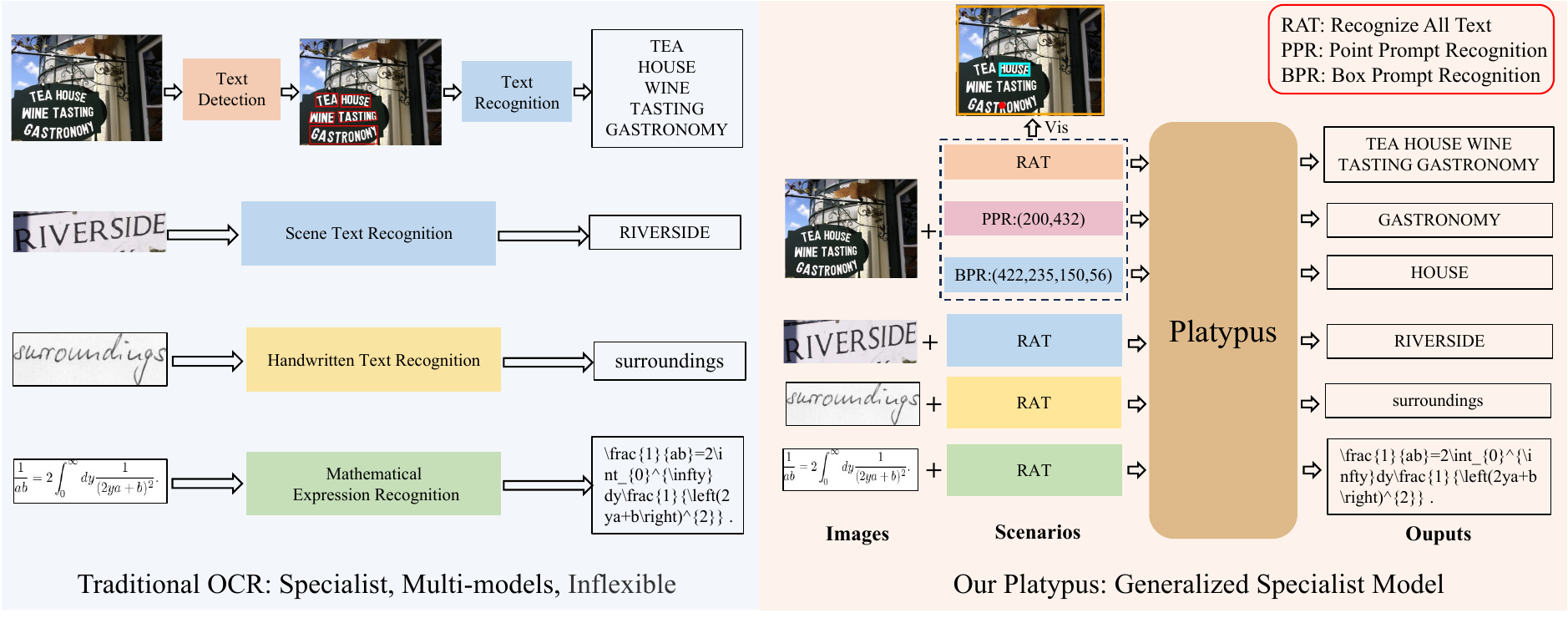}
    \caption{Comparison of our Platypus with previous OCR systems.}
    \label{fig:compare_with_ocr}
\end{figure}

Furthermore, Fig.~\ref{fig:compare_with_ocr} contrasts the integrated approach of Platypus with traditional OCR systems, underscoring its comprehensive and adaptable nature for handling a multitude of text reading tasks.

Our contributions are manifold:
\begin{itemize}
\item We introduce Platypus, a single unified architecture adept at various text reading tasks, serving as a versatile and unified text reading solution.
\item Platypus features a high degree of interactivity, providing users with the ability to specify areas for text recognition and select output granularity, enhancing usability and precision.
\item A comprehensive text reading dataset, Worms, is curated and presented, supporting the extensive training and evaluation of Platypus.
\item Platypus surpasses specialized text reading models and MLLMs in multiple text reading scenarios, establishing new state-of-the-art performances.
\end{itemize}

\section{Related Work}

\subsubsection{Scene Text Spotting Methods}
The goal of scene text spotting is to simultaneously detect and recognize all the text in an image in natural scene. Early end-to-end text spotting methods~\cite{2018cvpr_liu_fots,2019pami_liao_masktextspotterv2,2020cvpr_liu_abcnet,2019iccv_xing_charnet,2019iccv_feng_textdragonv1,2019iccv_qin_unconstrained,2021cvpr_liu_abcnetv2} concatenate module of text detection and recognition with modified ROI module. 
Recently, transformer-based methods~\cite{Zhang2022TESTR,Ye2022DeepSoloLT,Huang2023ESTextSpotterTB,Kil2023UNITS,Peng2021SPTS,Liu2023SPTSV2} achieve impressive achievement with their simple and efficient structures. 
DeepSolo~\cite{Ye2022DeepSoloLT} proposes learnable point queries to model text semantics and positions explicitly. 
The SPTS~\cite{Peng2021SPTS,Liu2023SPTSV2} series represent a text with a sequence combining center point and text.

\subsubsection{Scene Text Recognition Methods}
Scene text recognition has been extensively researched~\cite{Yao2012DetectingTO, yao2014strokelets, liao2022real, iccv2023lister}, initially focusing on rigid OCR systems for document analysis. The emergence of convolutional neural networks (CNNs) has shifted the focus towards models capable of handling complex variations in natural scenes. Researchers have incorporated recurrent neural networks (RNNs)~\cite{2016pami_shi_crnn,Hu2020GTC} and attention mechanisms~\cite{2018pami_shi_aster,Lee2016RecursiveRN,Cheng2017FocusingAT} to enhance performance. More recently, transformer-based models have been introduced, leveraging self-attention to capture long-range dependencies within text sequences~\cite{Fang2021ABINet,Bautista2022PARSeq,Li2021TrOCR,2020cvpr_yu_srn,Da2022LevOCR,Wang2022MultiGranularityPF,lister,mgpstrv2,DBLP:journals/pr/YangYLZB24}.

\subsubsection{Handwritten Text Recognition Methods}

The distinctiveness of handwritten text compared to printed text has necessitated specialized HTR models, usually requiring intricate preprocessing to normalize the wide range of handwriting styles and strokes~\cite{Bhunia2018HandwritingRI,Wang2019DecoupledAN,Luo2020LearnTA}. In the deep learning era, approaches such as Long Short-Term Memory (LSTM) networks have become prevalent, providing the ability to learn complex patterns in handwriting~\cite{Liu2022Persec,Yang2022DiG,Aberdam2020SeqCLR}.

\subsubsection{Mathematical Expression Recognition Methods}
MER poses unique challenges due to the spatial arrangements of symbols and the need to understand their semantic relationships. Approaches have extended beyond traditional symbol-based parsing~\cite{Alvaro2014RecognitionOO} to employ neural network architectures that can directly translate images of formulas into LaTeX representations~\cite{HME100K,Le2020RecognizingHM,Wu2020HandwrittenME}.

\subsubsection{Multi-modal Large Language Models}
MLLMs represent a significant advancement in text processing by incorporating additional modalities such as vision and language~\cite{Bai2023QwenVLAV,Feng2023UniDocAU,Ye2023mPLUGDocOwlMM,Li2023MonkeyIR,Liu2023LLaVA1.5}. These models have exhibited versatility across various applications. However, they often lack the fine-grained precision of specialized models~\cite{Shi2023ExploringOCR,Liu2023OnTH} and require substantial computational resources, limiting their practical deployment.

In summary, while text reading progress is notable, a holistic model for consistent performance across scenarios is needed. Platypus aims to provide a singular adaptive solution without the limitations of specialized and multimodal models.

\section{Methodology}

\subsection{Task Unification}
\label{sec:task_unification}

In addressing the diverse challenges of text reading from images, Platypus introduces a unified framework that encapsulates the various scenarios of text interpretation beyond traditional OCR boundaries. We categorize text reading tasks into four broad classes based on the source and presentation of the images.

\subsubsection{Category}
We define our categories as natural scene full images, document full images, cropped text, and cropped formulas. Natural scene and document full images often contain a heterogeneous mixture of text styles and layouts, necessitating a more versatile approach to recognition. For these images, we introduce three distinct recognition scenarios:
\begin{itemize}
\item \textit{RAT (Recognize All Text)}: This scenario involves recognizing all text within the entire image without any specific localization prompts.
\item \textit{PPR (Point Prompt Recognition)}: Text is recognized within a specified area surrounding a point prompt, allowing for targeted recognition of text.
\item \textit{BPR (Box Prompt Recognition)}: Text within a user-defined box area is recognized, which can be used to extract text from particular regions of interest.
\end{itemize}
For cropped text and formulas, the tasks involve direct interpretation of the text or mathematical expressions present in the images without the need for locational prompts, thus simplifying the recognition process.

\subsubsection{Writing Type}
Our model is trained to differentiate between two primary writing types: printed and handwritten. These inherently distinct types of text require unique recognition strategies, which Platypus accommodates seamlessly.

\subsubsection{Granularity}
We introduce granularity in recognition output to cater to the level of detail required by different text reading applications. This granularity is classified into two levels:
\begin{itemize}
\item \textit{Word-level}: At this level, the model focuses on recognizing individual words as the atomic units of text.
\item \textit{Line-level}: Here, the model recognizes entire lines of text, which may contain multiple words or even sentences.
\end{itemize}
For natural scene and document full image tasks, granularity is a critical parameter that determines the output's specificity, while for cropped text and formula tasks, the granularity is implicitly determined by the cropping, thus requiring no additional specification.

Through this structured categorization, Platypus demonstrates a versatile capability to interpret text from a variety of sources and formats, bridging the gap between specialized text reading models and the demands of real-world text recognition challenges.

\begin{figure}[t!]
    \centering
    \includegraphics[width=0.9\textwidth]{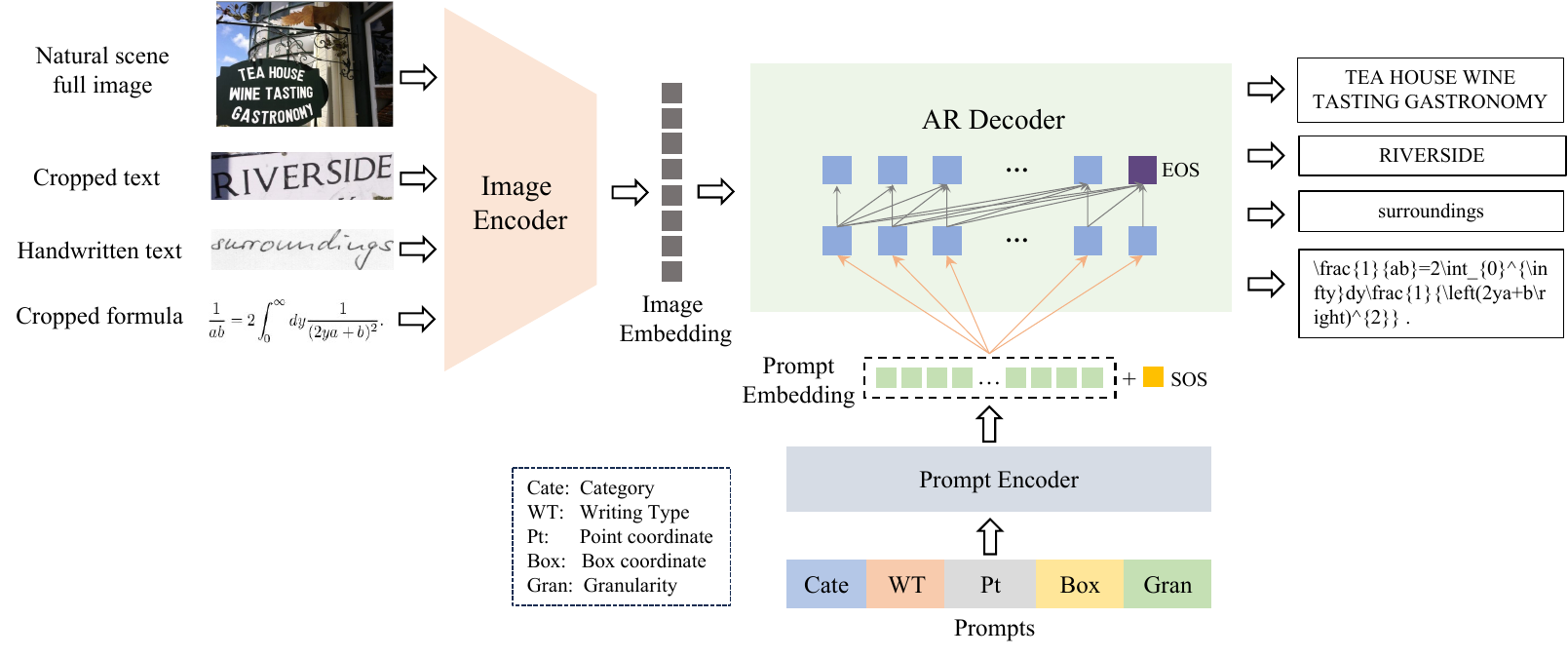}
    \caption{Overall architecture of our proposed Platypus model.}
    \label{fig:architecture}
\end{figure}

\subsection{Model Architecture}
\label{sec:model_architecture}
Our Platypus architecture is an encoder-decoder framework benefiting from the principles of the SAM~\cite{sam} model and incorporates a Prompt Encoder to encode various prompts. The overall architecture, illustrated in Fig.~\ref{fig:architecture}, combines these components into a cohesive system designed to robustly interpret text from various image forms.
\subsubsection{Image Encoder}
\label{sec:image_encoder}
The Image Encoder uses the pre-trained Swin-B Transformer~\cite{swin} on the ImageNet 22k dataset to extract multi-scale visual features. The extracted features are further enhanced by a Feature Pyramid Network (FPN)~\cite{fpn}, which provides a rich representation of the text at different scales and resolutions.

\subsubsection{Prompt Encoder}
\label{sec:prompt_encoder}

The Prompt Encoder is designed to handle the encoding of various prompts, which informs the model about the task category, writing type, output granularity, and the location information for the text to be recognized. We define the embedding for each category using a combination of position encoding and learned embeddings. For instance, the embedding for the task category is computed as follows:
\begin{equation}
\text{Embedding}_{\text{category}} = \text{PE} + E_{\text{category}}
\end{equation}
where $E_{\text{category}} \in \mathbb{R}^{5 \times d}$ represents the learned embeddings for four explicit task categories—natural scene full images, document full images, cropped text, and cropped formulas plus one additional embedding to handle cases where the image category is not specified. This setup enhances the model's flexibility, allowing it to perform robustly even when the category is unknown during inference. Similarly, writing type and granularity prompts include an unspecified category to maximize the model's adaptability across various text recognition scenarios. Follwing~\cite{sam}, the prompt embeddings for point and box are generated in a similar way, but for the box representation, we adopt a quadrilateral form with four points (top-left, top-right, bottom-right, bottom-left) to provide more precise positioning.

\subsubsection{Recognition Decoder}
\label{sec:recognition_decoder}

Inspired by the Transformer~\cite{transformer} architecture, our Recognition Decoder is an autoregressive module generating the output text sequence. It integrates the visual features from the Image Encoder and the prompt embeddings from the Prompt Encoder to produce the final text recognition result. It consists of 6 transformer layers with 8 heads each, both initialized randomly.

\subsection{Training Process}
\label{sec:training_process}

To effectively address the distinct challenges posed by different text reading scenarios, our training data is subdivided into two principal subsets. The first subset caters to full-image tasks such as natural scene full images and document full images, while the second subset is tailored for cropped-image tasks like cropped text and cropped formulas. The subsets not only differ in their image sizes but also utilize separate data loaders optimized for their specific needs.

The loss function for our model is an amalgamation of four individual loss components, each corresponding to separate recognition tasks. The full-image subset incurs three types of losses: $\mathcal{L}_{RAT-f}$ for the RAT task, $\mathcal{L}_{PPR-f}$ for the PPR task, and $\mathcal{L}_{BPR-f}$ for the BPR task. In contrast, the cropped-image subset is associated with a single loss, $\mathcal{L}_{RAT-c}$, related solely to the RAT task. Each of these losses is weighted equally and set to 1 for simplicity.

\begin{equation}
\mathcal{L}_{total} = \lambda_1 \mathcal{L}_{RAT-f} + \lambda_2 \mathcal{L}_{PPR-f} + \lambda_3 \mathcal{L}_{BPR-f} + \lambda_4 \mathcal{L}_{RAT-c},
\end{equation}

Each of these losses is computed using standard cross-entropy loss, which measures the predicted probabilities against the true labels of the text across our diverse dataset.

\subsection{Inference Process}
\label{sec:inference_process}

During inference, Platypus leverages a streamlined approach where the model generates predictions based on the input scene category and specific prompts provided. This enables the model to accurately interpret and recognize text from a wide range of input images without the direct need for detection or segmentation models.

\section{Experiments}
In this section, we conduct both qualitative and quantitative experiments on standard benchmarks of Scene Text Spotting (STS), Scene Text Recognition (STR), Handwritten Text Recognition (HTR), and Mathematical Expression Recognition (MER), to verify the effectiveness and advantages of Platypus.

\subsection{Implementation Details}

To train our Platypus for robust text reading across diverse scenarios, we meticulously prepared our dataset and conducted a two-phase training process.

\textbf{Data Preparation} As discussed in Sec.~\ref{sec:training_process}, we partition our training data into two subsets to accommodate the distinct dimensions of full-image and cropped-image tasks. Full-size images, encompassing natural scenes and documents, are resized to 1024 pixels on their longer edge. For cropped images, including text snippets and formulas, we resize the longer size to 768 pixels. 

\textbf{Pre-training} The initial phase focused on pre-training with full image data, which presents complex, large-scale recognition challenges. For the RAT scenario, we set the point prompt to [0,0] and defined the box prompt as the full image size. For PPR, points are uniformly sampled within the text bounding boxes, while for BPR, ground-truth boundary boxes are perturbed with noise (10\% of the box size, capped at 20 pixels). Besides, during each forward pass, multiple points and boxes are selected from a single image, capped at a maximum of eight prompts to allow the model to thoroughly learn the contextual relationships of text entities in the image. And granularity is selected randomly when both word and line annotations are available, guiding the corresponding ground truth selection. We employ a batch size of 2, augmented with instance-aware random cropping, rotations between -90\textdegree and 90\textdegree, random scaling, and color jittering. AdamW optimizer is utilized in the pre-training stage, with an initial learning rate of $5e^{-4}$, for a total of 1,000k steps, implementing a warm-up schedule for the first 5k steps and then linearly decaying the learning rate to zero.

\textbf{Joint Training} After the pre-training stage, we introduce cropped-image data into training. Separate data loaders are maintained for each subset, and batches alternated between full-image and cropped-image data to ensure a diversified learning experience. The data preparation and training techniques for the full-image subset remain consistent with the pre-training phase. For the cropped-image subset, recognizing the granularity of text is not necessary, as these images are pre-cropped to contain individual words or lines of text. Only the RAT scenario is applicable, with point prompts set to [0,0] and box prompts designed to cover the entire image size. A larger batch size of 16 is utilized for cropped images, and we apply common text image augmentation methods such as perspective and affine distortions, blurring, noise, and rotation to simulate various environmental conditions. The joint training continued with the AdamW optimizer, for 500k steps, using an initial learning rate of $3e^{-4}$ and the same learning rate schedule as established in the pre-training phase. All training is conducted on eight NVIDIA A100 GPUs.

\textbf{Inference} 
For full-image tasks, such as Scene Text Spotting (STS), we resize the longer size of image to 1024 and keep the aspect ratio. In the RAT recognition scenario, we adopt box prompt encompassing the entire image for evaluation. For the PPR scenario, the center point of each polygon annotation is adopted, and for the BPR scenario, the quadrilateral bounding box of groundtruth is used. For cropped-image tasks, such as Scene Text Recognition (STR), Handwritten Text Recognition (HTR), and Mathematical Expression Recognition (MER), which only entail the RAT scenario, we resize the longer size of image to 768, while keeping its aspect ratio. All evaluations are performed on one NVIDIA V100 GPU.

\subsection{Comprehensive Text Reading Datasets (Worms)}

The proposed model is trained in a unified way, thus, training data for STR, HTR, MER, and STS are combined for training. We collected and organized data from a multitude of public resources to build comprehensive text reading datasets, termed \textbf{Worms}. These datasets serve as the primary training set, akin to the favored diet of the Platypus in our model's training framework. Comprehensive details about the Worms dataset, including the specifics of each subset used, can be found in the Appendix.

\subsection{Benchmarks and Evaluation Protocols}
\noindent\textbf{Scene Text Spotting (STS)}
To assess the ability scene text spotting of different methods, we evaluate on four popular scene text datasets, Total-Text~\cite{totaltext}, ICDAR 2013 (IC13)~\cite{IC13}, ICDAR 2015 (IC15)~\cite{IC15}, and CTW1500~\cite{ctw1500}.

\noindent\textbf{Scene Text Recognition (STR)}
Experiments are conducted on six standard Latin scene text benchmarks, including 3 regular text datasets (IC13~\cite{IC13}, SVT~\cite{SVT}, IIIT~\cite{IIIT}) and 3 irregular ones (IC15~\cite{IC15}, SVTP~\cite{SVTP}, CUTE~\cite{CUTE}). 

\noindent\textbf{Handwritten Text Recognition (HTR)}
To validate the performance on HTR, we also conduct experiments on handwritten text recognition datasets with English (IAM~\cite{IAM} and CVL~\cite{CVL}) and French (RIMES~\cite{RIMES}). In accordance with the methodologies employed in DiG~\cite{Yang2022DiG}, test sets of CVL and IAM include 12,012 and 13,752 cropped images respectively. For the RIMES dataset, 7,776 cropped images are used for testing.

\noindent\textbf{Mathematical Expression Recognition (MER)}
For task of Mathematical Expression Recognition (MER), we use the benchmark of Latex-OCR\footnote{https://github.com/lukas-blecher/LaTeX-OCR} with cropped images of print formula. For convenience, we randomly select 1000 images of the test set for evaluation on MER.

\begin{table}[t!]
  \centering
  \caption{Comparison between Platypus and SOTA models on different text reading tasks, including MLLMs. Specialist models are limited to particular tasks, while Platypus, similar to MLLMs, is capable of accommodating multiple tasks.}
  \begin{tabular}{c|c|c|c|c|c|c}
    \hline
    \multicolumn{2}{c|}{SOTA Methods} & STS & STR & HTR & MER & Params (M) \\
    \hline
    \multirow{3}{*}{STS Models} & SwinTextSpotter~\cite{Huang2022SwinTextSpotterST} & \greentick & \redcross & \redcross & \redcross & - \\
    & SPTS~\cite{Peng2021SPTS} & \greentick & \redcross & \redcross & \redcross & 36.5 \\
    & DeepSolo~\cite{Ye2022DeepSoloLT} & \greentick & \redcross & \redcross & \redcross & 42.5 \\
    \hline
    \multirow{3}{*}{STR Models} & MGP-STR~\cite{Wang2022MultiGranularityPF} & \redcross & \greentick & \greentick & \redcross & 148.0 \\
    & ABINet~\cite{Fang2021ABINet} & \redcross & \greentick & \greentick & \redcross & 36.7 \\
    & PARSeq~\cite{Bautista2022PARSeq} & \redcross & \greentick & \greentick & \redcross & 23.8 \\
    \hline
    \multirow{3}{*}{HTR Models} & SeqCLR~\cite{Aberdam2020SeqCLR}  & \redcross & \greentick & \greentick & \redcross & - \\
    & PerSec-ViT~\cite{Liu2022Persec} & \redcross & \greentick & \greentick & \redcross & - \\
    & DiG-ViT-Base~\cite{Yang2022DiG} & \redcross & \greentick & \greentick & \redcross & 52 \\
    \hline
    MER Models & LaTeX-OCR  & \redcross & \redcross & \redcross & \greentick & 25.5 \\
    \hline
    \multirow{3}{*}{MLLMs} & mPLUG-Owl2~\cite{Ye2023mPLUGDocOwlMM}  & \greentick & \greentick & \greentick & \greentick & 8200 \\
    & LLaVA1.5-7B~\cite{Liu2023LLaVA1.5} & \greentick & \greentick & \greentick & \greentick  & 7000 \\
    & GPT-4V~\cite{2023GPT4VisionSC} & \greentick & \greentick & \greentick & \greentick & - \\
    \hline
    \multicolumn{2}{c|}{Ours} & \greentick & \greentick & \greentick & \greentick & 118.8 \\
    \hline
    \end{tabular}
    \label{tab:overall evaluation2}
\end{table}

\subsubsection{Evaluation Metrics}
\label{sec:metric}
For benchmarks of STS, end-to-end text spotting methods are evaluated with H-mean of recall and precision on recognition of given lexicons. 
For MLLMs, the predictions and ground truths (GT) are split with spaces and then evaluated with precision and recall according to evaluation metric as defined in~\cite{Shi2023ExploringOCR}. Precision stands for the percentages of correctly identified words to those generated by MLLMs, while recall is the ratio of correctly identified words to the total number of GT words. 
Thus H-mean is calculated through recall and precision. 

For evaluation of the proposed Platypus on STS benchmarks, we calculated H-mean as the same as MLLMs, given the recognition scenarios such as RAT, PPR and BPR. While for benchmarks of STR and HTR, we evaluate performance with word accuracy ignoring case and symbols (WAICS)~\cite{Union14M} as metric. 
For MER, we evaluate the performance using the Word Error Rate (WER) metric and Character Error Rate (CER).

\subsection{Comparison with State-of-the-Art Specialists and MLLMs}
\label{sec:compare_sota}

\subsubsection{Overall Capability Comparisons on Extensive Text Reading Benchmarks}
We firstly make capability comparisons between our models and state-of-the-art (SOTA) models on different text reading tasks, including MLLMs. As can be concluded in Tab.~\ref{tab:overall evaluation2}, previous specialist SOTA models can only prevail on specific text reading tasks. While the MLLMs and our Platypus are able to compete on various text reading tasks as generalist models.

\begin{table}[t!]
  \centering
  \caption{Performance comparison between our method and SOTA text spotting models. For GPT-4V, we filter images with no results to evaluate the rest images marked with \textit{valid}, and evaluation on all images is marked with \textit{all}. }
  \begin{tabular}{c|c|cc|cc|cc|cc}
    \hline
    SOTA Methods & Scenarios & \multicolumn{2}{|c|}{IC13} & \multicolumn{2}{|c|}{IC15} & \multicolumn{2}{|c|}{TotalText} & \multicolumn{2}{|c}{CTW1500} \\
    \hline
    Lexicon & - & G & None & G & None & None & None & None & None \\
    \hline
    Granularity & - & word & line & word & line & word & line & word & line \\
    \hline
    ABCNet v2~\cite{2021cvpr_liu_abcnetv2} & Spotting & - & - & 73.0 & - & 70.4 & - & - & 57.5 \\
    MANGO~\cite{Qiao2020MANGO} & Spotting & 88.7 & - & 73.9 & - & 72.9 & - & - & 58.9 \\
    SwinTextSpotter~\cite{Huang2022SwinTextSpotterST} & Spotting & - & - & 70.5 & - & 74.3 & - & - & 51.8 \\
    TESTR~\cite{Zhang2022TESTR} & Spotting & - & - & 73.6 & - & 73.3 & - & - & 56.0 \\
    SPTS~\cite{Peng2021SPTS} & Spotting & 88.5 & - & 65.8 & - & 74.2 & - & - & 63.6 \\
    ESTextSpotter~\cite{Huang2023ESTextSpotterTB} & Spotting & - & - & 78.1 & - & 80.8 & - & - & 66.0 \\
    DeepSolo~\cite{Ye2022DeepSoloLT} & Spotting & 90.1 & - & 79.1 & - & 82.5 & - & - & 64.2 \\
    GPT-4V~\cite{2023GPT4VisionSC}(\textit{all}) & RAT & 88.3 & 52.1 & 34.4 & 20.6 & 70.7 & 45.2 & 75.7 & 51.7 \\
    GPT-4V~\cite{2023GPT4VisionSC}(\textit{valid}) & RAT & 88.8 & 52.1 & 38.8 & 20.1 & 71.1 & 45.5 & 75.9 & 51.9 \\
    \hline
    \multirow{3}{*}{Ours} & RAT & 92.7 & 83.8 & 68.3 & 63.1 & 78.9 & 58.0 & 80.9 & 73.5 \\
                          & PPR   & 94.6 & 85.9 & 85.7 & 78.7 & 88.3 & 76.9 & 86.6 & 75.6 \\
                          & BPR     & \textbf{96.9} & \textbf{86.0} & \textbf{88.2} & \textbf{77.9} & \textbf{93.0} & \textbf{77.4} & \textbf{91.9} & \textbf{76.1} \\
    \hline
    \end{tabular}
    \label{tab:STS2}
\end{table}

\subsubsection{Comparison on Scene Text Spotting (STS)}
To further demonstrate the performance on natural scene full images, we perform evaluation on Scene Text Spotting benchmarks, in comparison with MLLMs and previous SOTA scene text spotters. All the evaluations of MLLMs and the proposed Platypus model are under metric defined in Sec.~\ref{sec:metric} and the comparison results are displayed in Tab.~\ref{tab:STS2}. For comparison with different prompts on Platypus, box prompt (i.e. BPR) and point prompt (i.e. PPR) with clear indication where text may be located, perform better than fullbox prompt (i.e. RAT). Compared with specialist models on STS, which are usually fine-tuned on the specific dataset and support only one granularity, Platypus can read text with word-level and line-level granularity. Besides, the performance of Platypus in PPR and BPR settings is better than that of SOTA on four benchmarks, and Platypus with RAT scenario is also comparable with specialist models on STS. When in comparison with MLLM GPT-4V, the proposed Platypus surpasses it with a large gap even in the RAT setting.

\begin{table}[h]
  \small
  \centering
  \caption{Performance comparison between our method and SOTA scene text recognition models. Note that the results of MLLMs with $\dagger$ are obtained from \cite{Liu2023OnTH}.}
  \begin{tabular}{c|c|cccccc|c}
    \hline
    \multicolumn{2}{c|}{Methods} & IIIT5k & SVT & IC13 & IC15 & SVTP & CUTE80 & Average\\
    \hline
    \multirow{5}{*}{Specialist Models} & ViTSTR\cite{Atienza2021VisionTF} & 88.4 & 87.7 & 93.2 & 78.5 & 81.8 & 81.3 & 85.6 \\
    & SRN~\cite{2020cvpr_yu_srn} & 94.8 & 91.5 & 95.5 &  82.7 & 85.1 & 87.8 & 90.4 \\
    & ABINet~\cite{Fang2021ABINet} & 96.2 & 93.5 & 97.4 & 86.0 & 89.3 & 89.2 & 92.6 \\
    & TrOCR~\cite{Li2021TrOCR} & 91.0 & 93.2 & 98.3 & 84.0 & 91.0 & 89.6 & 90.2 \\
    & MGP-STR~\cite{Wang2022MultiGranularityPF} & 96.4 & 97.3 & 94.7 & 87.2 & 91.0 & 90.3 & 93.3 \\
    & PARSeq~\cite{Bautista2022PARSeq} & 99.1 & 97.9 & 98.3 & 90.7 & \textbf{95.7} & 98.3 & 96.4 \\
    \hline
    \multirow{6}{*}{Generalist Models} & mPLUG-Owl2$\dagger$ & 80.9 & 69.6 & 79.8 & 53.9 & 53.5 & 74.8 & 70.3 \\
    & LLaVA1.5-7B$\dagger$ & 84.2 & 85.7 & 86.4 & 71.9 & 79.8 & 82.7 & 81.0 \\
    & UniDoc$\dagger$ & 91.9 & 89.2 & 90.9 & 78.0 & 80.3 & 88.2 & 86.9 \\
    & Monkey$\dagger$ & 83.7 & 75.1 & 85.4 & 53.4 & 58.4 & 73.9 & 72.9 \\
    & GPT-4V~\cite{2023GPT4VisionSC}(\textit{all}) & 37.3 & 66.3 & 66.1 & 39.0 & 52.4 & 68.3 & 46.2 \\
    & GPT-4V~\cite{2023GPT4VisionSC}(\textit{valid}) & 69.4 & 91.0 & 90.9 & 70.8 & 86.9 & 93.4 & 76.7 \\
    \hline
    \multicolumn{2}{c|}{Ours} & \textbf{99.1} & \textbf{98.1} & \textbf{98.5} & \textbf{90.7} & 95.1 & \textbf{98.9} & \textbf{96.4} \\
    \hline
    \end{tabular}
    \label{tab:STR}
\end{table}

\subsubsection{Comparison on Scene Text Recognition (STR)}
We also compare our proposed Platypus with SOTA models and MLLMs on STR, and the results on 6 standard benchmarks are summarized in Tab.~\ref{tab:STR}. 
As can be seen from the results, though Platypus is a unified model for 4 tasks, Platypus can also achieve SOTA results surpassing not only generalist models such as GPT-4V and Monkey but also STR specialist models (i.e., ABINet~\cite{Fang2021ABINet}, MGP-STR~\cite{Wang2022MultiGranularityPF} and PARSeq~\cite{Bautista2022PARSeq}) on fine-grained task as STR. 

\subsubsection{Comparison on Handwritten Text Recognition (HTR)}
To validate the adaptability of Platypus on various types such as handwritten text, we conduct evaluation on HTR benchmarks, comparing with previous SOTA methods and MLLMs (i.e., GPT-4V). 
As depicted in Tab.~\ref{tab:HTR}, GPT-4V suffers from poor accuracy on 3 Latin benchmarks of HTR, 
indicating poor adaptability to handwritten text. The proposed Platypus instead sets a new SOTA on three benchmarks, and the performance gains are 8.2\%, 0.1\% and 1.2\%, compared with the previous SOTA on IAM, CVL and RIMES respectively. 

\begin{table}[ht]
  \small
  \begin{minipage}{0.47\linewidth}
      \centering
      \small
      \caption{Performance comparison between our method and SOTA handwritten text recognition models.}
      \begin{tabular}{c|ccc}
        \hline
        Methods & IAM & CVL & RIMES \\
        \hline
        SeqCLR~\cite{Aberdam2020SeqCLR}                   & 79.9 & 77.8 & 92.4 \\
        PerSec-ViT~\cite{Liu2022Persec}               & 83.7 & 82.9 & - \\
        DiG-ViT-Base~\cite{Yang2022DiG}             & 87.0 & 91.3 & - \\
        GPT-4V~\cite{2023GPT4VisionSC}(\textit{all})                & 40.4 & 26.9 & 15.9 \\
        GPT-4V~\cite{2023GPT4VisionSC}(\textit{valid})              & 50.7 & 38.3 & 26.0 \\
        \hline
        Ours                     & \textbf{96.4} & \textbf{91.4} & \textbf{93.6} \\
        \hline
        \end{tabular}
        \label{tab:HTR}
    \end{minipage}
    \hfill
    \begin{minipage}{0.47\linewidth}
      \centering
      \caption{Performance comparison between Platypus and SOTA mathematical expression recognition models.}
      \begin{tabular}{c|cc}
        \hline
        Methods & \multicolumn{2}{|c}{LaTeX-OCR} \\
        & CER($\downarrow$) & WER($\downarrow$) \\
        \hline
        LaTeX-OCR & 8.7 & 9.6 \\
        GPT-4V~\cite{2023GPT4VisionSC}(\textit{all}) & 47.2 & 50.1 \\
        GPT-4V~\cite{2023GPT4VisionSC}(\textit{valid}) & 47.8 & 50.7 \\
        \hline
        Ours & \textbf{7.2} & \textbf{7.8} \\
        \hline
        \end{tabular}
        \label{tab:MER}        
    \end{minipage}
\end{table}

\subsubsection{Comparison on Mathematical Expression Recognition (MER)}
We also compare Platypus with MLLMs and open source formula recognition method on MER benchmark LaTeX-OCR. As seen in Tab.~\ref{tab:MER}, on structured text as formula, GPT-4V has poor adaptation performance. While the performance of Platypus can also surpass LaTeX-OCR for 1.5\% and 1.8\% in metric CER and WER respectively.

\subsubsection{Comparisons of efficiency (FPS)}
The comparisons of inference speed (evaluated on a single V100 GPU) are depicted in~Tab.~\ref{tab:Efficiency evaluation2}. As can be observed, Platypus exhibits superior inference speed, surpassing MLLMs (\eg mPLUG-Owl2) and achieving higher efficiency than specialist models on STS, HTR and MER tasks. In particular, Platypus even runs $\times3$ faster than SPTS, mainly attributed to its efficient single-pass decoding.

\subsection{Analyses}

\subsubsection{Exploring the Effectiveness of Prompts}
\label{subsec:prompts_effectiveness}


The efficacy of prompts in guiding text recognition was extensively examined. In the STS benchmark, point, box, and granularity prompts were invaluable for full-image tasks. For cropped-image tasks, we assessed category and writing type prompts across four scenarios: no prompts, only category, only writing type, and both combined (see Tab.~\ref{tab:prompt_effectiveness}). The combined prompts led to the best recognition accuracy, especially in HTR and MER tasks, with MER being more sensitive to category prompts. STR showed reasonable accuracy even without prompts, indicating robustness. These findings highlight the adaptive utility of prompts in various recognition contexts.

\subsubsection{Evaluating Different Pre-training Approaches}
\label{subsec:pretraining_approaches}

We explored three pre-training strategies, as shown in Tab.~\ref{tab:pretraining_comparison}. The methods included (1) joint training without pre-training, (2) pre-training on cropped-image data followed by joint training, and (3) pre-training on full-image data followed by joint training with cropped-image data. Each method underwent a 200k-step training regimen. Evaluation on selected subsets from STS, STR, HTR, and MER benchmarks revealed significant accuracy gains when full-image pre-training was applied, especially notable in word-level IC15 and line-level CTW1500 for STS. Models pre-trained on full images showed marked improvements in STR, HTR, and MER tasks, affirming the benefit of this approach. Notably, the lack of pre-training led to substantial performance deficits, underscoring its importance.

\subsubsection{Impact of Image Size}
\label{subsec:image_size_impact}

To ascertain the influence of image resolution on text recognition, we conducted experiments with two distinct image size settings. For full-image tasks, we compared the performance of models trained with a maximum edge length of 768 pixels against those trained with 1024 pixels, while for cropped-image tasks, we contrasted the sizes of 512 and 768 pixels. Following a pre-training phase of 1,000k steps, both models underwent 500k steps of joint training. The outcome, presented in Tab.~\ref{tab:comparison_img_size}, indicates an improvement in accuracy across all tasks when larger image sizes are used, with full-image tasks benefiting the most. This confirms our hypothesis that higher resolutions significantly aid in recognizing text within larger images, with a less pronounced but still considerable effect observed for cropped-image tasks.

\begin{table}[t!]
  \centering
  \small
  \caption{The effectiveness of category prompt and writing type prompt.}
      \begin{tabular}{c|c|ccc|c|c}
        \hline
        \multicolumn{2}{c|}{Prompt} & \multicolumn{3}{|c|}{STR(ACC $\uparrow$)} & HTR(ACC $\uparrow$) & MER(WER $\downarrow$) \\
        \cline{1-2}
        Category & Writing type & IIIT5k & IC15  & CUTE80 & IAM & LaTeX-OCR \\
        \hline
        \redcross & \redcross & 98.8 & 90.5 & 98.5 & 88.6 & 8.5 \\
        \hline
        \greentick & \redcross & 98.8 & 90.5 & 98.6 & 94.1 & 7.9 \\
        \hline
        \redcross & \greentick & 98.8 & 90.7 & 98.3 & 96.2 & 8.5 \\
        \hline
        \greentick & \greentick & 99.1 & 90.7 & 98.9 & 96.4 & 7.8 \\
        \hline
        \end{tabular}
        \label{tab:prompt_effectiveness}
\end{table}

\begin{table}[t!]
  \centering
  \caption{Performance comparison between different pre-training strategies.}
  \resizebox{\textwidth}{!}{
      \begin{tabular}{c|cc|ccc|c|c}
        \hline
        \multirow{2}{*}{Strategy} & \multicolumn{2}{|c|}{STS(H-mean $\uparrow$)} & \multicolumn{3}{|c|}{STR(ACC $\uparrow$)} & HTR(ACC $\uparrow$) & MER(WER $\downarrow$) \\
                                      & IC15 & CTW1500 & IIIT5k & IC15  & CUTE80 & IAM & LaTeX-OCR \\
        \hline
        w/o pre-training & 0.1 & 0.1 & 2.3 & 2.6 & 1.1 & 4.1 & 98.7 \\
        \hline
        cropped image pre-training & 61.2 & 63.0 & 98.5 & 88.1 & 96.5 & 92.3 & 8.7 \\
        \hline
        full image pre-training & 86.7 & 76.1 & 98.7 & 90.2 & 97.5 & 95.5 & 8.3 \\
        \hline
        \end{tabular}
        \label{tab:pretraining_comparison}
}
\end{table}

\begin{table}[t!]
  \centering
  \caption{Performance comparison between different training image sizes.}
  \resizebox{\textwidth}{!}{
      \begin{tabular}{c|c|cc|ccc|c|c}
        \hline
        \multirow{2}{*}{full-img size} & \multirow{2}{*}{cropped-img size} & \multicolumn{2}{|c|}{STS(H-mean $\uparrow$)} & \multicolumn{3}{|c|}{STR(ACC $\uparrow$)} & HTR(ACC $\uparrow$) & MER(WER $\downarrow$) \\
                                      & & IC15 & CTW1500 & IIIT5k & IC15  & CUTE80 & IAM & LaTeX-OCR \\
        \hline
        1024 & 768 & 88.2 & 76.1 & 99.1 & 90.7 & 98.9 & 96.4 & 7.8 \\
        \hline
        768 & 512 & 82.4 & 75.4 & 98.9 & 90.5 & 98.9 & 96.3 & 8.3 \\
        \hline
        \end{tabular}
        \label{tab:comparison_img_size}
}
\end{table}

\begin{table}[h]
\begin{minipage}{0.48\linewidth} 
  \centering
  \caption{\small{Comparison of FPS.}} \label{tab:Efficiency evaluation2}
  \small
  \scalebox{0.7}{
  \begin{tabular}{c|c|c|c|c|c}
    \hline
    \multicolumn{2}{c|}{\multirow{2}{*}{\textbf{SOTA Methods}}} & \textbf{STS} & \textbf{STR} & \textbf{HTR} & \textbf{MER}  \\
    \cline{3-6}
    \multicolumn{2}{c|}{} & TotalText & IC13 & IAM & Latex-OCR  \\
    \hline
    STS  & SPTS & 0.3 & - & - & - \\
    \hline
    STR  & PARSeq & - & 25.9 & - & - \\
    \hline
    HTR  & DiG-ViT-Base & - & - & 4.6 & - \\
    \hline
    MER  & LaTeX-OCR  & - & - & - & 1.3 \\
    \hline
    \multirow{2}{*}{MLLMs} & mPLUG-Owl2  & 0.4 & 2.8 & 2.2 & 0.4 \\
    \cline{2-6}
    & GPT-4V & 0.3 & 0.7 & 0.8 & 0.4 \\
    \hline
    \multicolumn{2}{c|}{Ours} & \textbf{1.0} & 11.9 & \textbf{13.0} & \textbf{2.1} \\
    \hline
    \end{tabular}}
\end{minipage}
\hfill
\begin{minipage}{0.48\linewidth}
  \small
  \centering
  \caption{Performance comparison on CAT.}
  \scriptsize
  \scalebox{0.85}{
  \begin{tabular}{c|c|ccc}
    \hline
    Methods & Scenarios & Recall & Precision & H-mean \\
    \hline
    PaddleOCR & Spotting & 43.7 & 62.6 & 51.4 \\
    \hline
    EasyOCR & Spotting & 38.5 & 48.5 & 42.9 \\
    \hline
    GPT-4V~\cite{2023GPT4VisionSC} & RAT & 80.1 & 85.1 & 82.5 \\
    \hline
    Qwen-VL-Plus~\cite{Bai2023QwenVLAV} & RAT & 69.2 & 81.9 & 75.0 \\
    \hline
    \multirow{3}{*}{Ours} & RAT & 72.7 & 84.4 & 78.1 \\
                          & PPR & 84.3 & 84.3 & 84.3 \\
                          & BPR & \textbf{89.8} & \textbf{89.8} & \textbf{89.8} \\
    \hline
    \end{tabular}}
    \label{tab:cat benchmark}
\end{minipage}
\end{table} 

\begin{figure}[ht]
    \centering
    \includegraphics[width=0.98\textwidth]{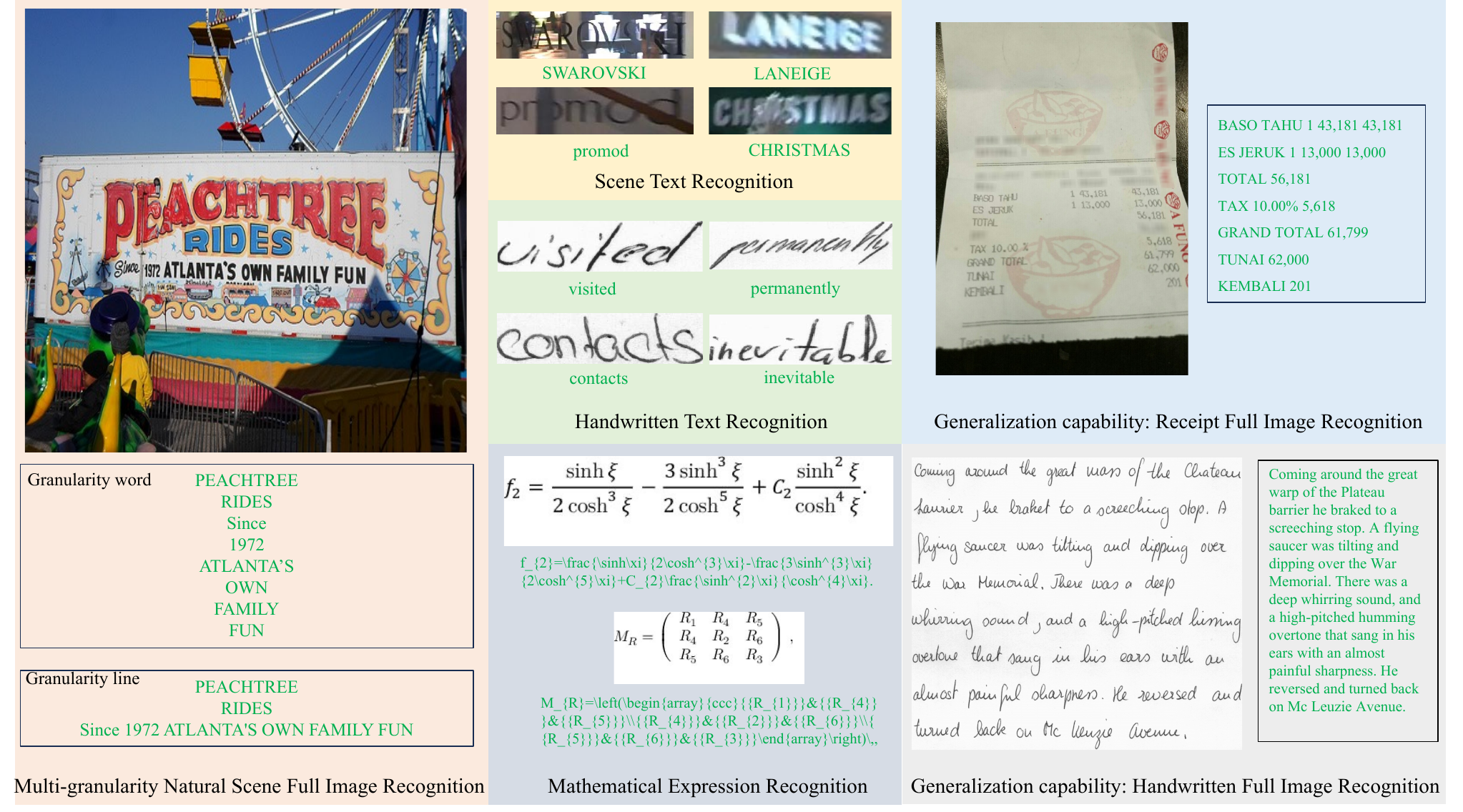}
    \caption{Qualitative results of Platypus. All images are derived from public datasets.}
    \label{fig:qualitative}
\end{figure}


\subsubsection{Comparisons on Curated Artistic Text (CAT) Benchmark} 100 images from LAION-5B~\cite{Schuhmann2022LAION5BAO} were curated and annotated to evaluate text reading pipelines on multi-orientation, occluded, overlapped, and artistic text. We make comparisons among proposed Platypus, MLLMs (GPT-4V and Qwen-VL-Plus), and open source OCR pipelines (PaddleOCR\footnote{https://github.com/PaddlePaddle/PaddleOCR} and EasyOCR\footnote{https://github.com/JaidedAI/EasyOCR}) on this benchmark. Note that, the results of MLLMs are obtained from the APIs, and the results of open-source OCR pipelines are produced from the officially released codes and models. As displayed in Tab.~\ref{tab:cat benchmark}, Platypus outperforms PaddleOCR and EasyOCR by 26.7\% and 35.2\% H-mean, respectively. Compared to GPT-4V on RAT, Platypus achieves comparable performance and better results on PPR and BPR scenarios. The qualitative comparisons are shown in Fig.~\ref{fig:laionocr}. We can conclude through the results that the proposed Platypus has better generalization ability on CAT benchmark with multi-orientation, artistic shape and overlapped issues, compared to OCR pipelines and even MLLMs. 

\subsubsection{Qualitative Results}
We display some qualitative results of Platypus in Fig.~\ref{fig:qualitative}. The qualitative results further demonstrate the ability of Platypus on multi-granularity natural scene full image recognition, scene text and handwritten text recognition, mathematical expression recognition, and generalized capability on unseen images such as receipt full images and handwritten full images.

\section{Conclusion}

In summary, Platypus unifies text reading across various forms and complexities, achieving SOTA performance and outperforming specialized and multimodal models. The interactive prompt mechanism enhances precision, paving the way for future improvements. Although our evaluations focused on English text, there is potential for multi-language expansion. Future efforts will target linguistic diversity and refine the model's real-world text interpretation capabilities.

%
%
\bibliographystyle{splncs04}
\bibliography{main}

\clearpage
\appendix
\section{Appendix}
\subsection{Comprehensive Text Reading Datasets (Worms) Composition}
The following tables provide detailed statistics of the Worms training set. The dataset includes diverse subsets to cover various text reading tasks comprehensively.

\begin{table}[ht]
  \centering
  \caption{The statistics of the training set of our proposed comprehensive text reading datasets (Worms). Dataset marked with an * suggests that the annotations are inexact, potentially representing pseudo-labels generated through model-based pre-annotation. In the "Granularity" column, the presence of a blue star \textcolor{blue}{$\star$} signifies that the original dataset lacked annotations, which we have supplemented. Synth-arxiv with $\dagger$ means synthetic formula data generated by ourselves. }
  \resizebox{\textwidth}{!}{
      \begin{tabular}{c|c|c|c|cc|c}
        \hline
        \multirow{2}{*}{Dataset} & \multirow{2}{*}{Subset} & \multirow{2}{*}{Category} & Writing & \multicolumn{2}{|c|}{Granularity} & \multirow{2}{*}{Number} \\
        \cline{5-6}
        & & & Type & word & line \\
        \hline
        ICDAR2013~\cite{IC13} & train & \multirow{12}{*}{Natural scene full image} & \multirow{19}{*}{printed} & \greentick & \textcolor{blue}{$\star$} & 229 \\
        \cline{1-2}
        \cline{5-7}
        ICDAR2015~\cite{IC15} & train & & & \greentick & \textcolor{blue}{$\star$} & 1K \\
        \cline{1-2}
        \cline{5-7}
        CTW1500~\cite{ctw1500} & train & & & \textcolor{blue}{$\star$} & \greentick & 1K \\
        \cline{1-2}
        \cline{5-7}
        TotalText~\cite{totaltext} & train & & & \greentick & \textcolor{blue}{$\star$} & 1.3K \\
        \cline{1-2}
        \cline{5-7}
        HierText~\cite{Long2022HierText} & trainval & & & \greentick & \greentick & 10K \\
        \cline{1-2}
        \cline{5-7}
        TextOCR~\cite{TextOCR} & train & & & \greentick & \textcolor{blue}{$\star$} & 25K \\
        \cline{1-2}
        \cline{5-7}
        Open Images V5 Text~\cite{OpenVINO} & trainval & & & \greentick & \redcross & 208K \\
        \cline{1-2}
        \cline{5-7}
        Uber-Text~\cite{Uber} & trainval & & & \redcross & \greentick & 83K \\
        \cline{1-2}
        \cline{5-7}
        COCO-Text~\cite{COCO} & trainval & & & \greentick & \redcross & 54K \\
        \cline{1-2}
        \cline{5-7}
        Curved SynthText~\cite{2021cvpr_liu_abcnetv2} & train & & & \greentick & \redcross & 149K \\
        \cline{1-2}
        \cline{5-7}
        MLT2017~\cite{MLT17} & train & & & \greentick & \redcross & 10K \\
        \cline{1-2}
        \cline{5-7}
        LAION-OCR~\cite{Schuhmann2022LAION5BAO}\raisebox{-0.5ex}{\textsuperscript{*}} & train & & & \greentick & \redcross & 2.5M \\
        \cline{1-3}
        \cline{5-7}
        PubLayNet~\cite{Zhong2019PubLayNetLD}\raisebox{-0.5ex}{\textsuperscript{*}} & trainval & Document full image & & \greentick & \greentick & 352K \\
        \cline{1-3}
        \cline{5-7}
        MJSynth~\cite{MJ1} & train & \multirow{9}{*}{Cropped text} & & \multicolumn{2}{|c|}{\multirow{13}{*}{none}} & 8.9M \\
        \cline{1-2}
        \cline{7-7}
        SynthText~\cite{syntext800k} & train & & & & & 7.3M \\
        \cline{1-2}
        \cline{7-7}
        SynthAdd\tablefootnote{From https://mmocr.readthedocs.io/en/v0.6.3/datasets/recog.html\#synthadd.} & train & & & & & 1.2M \\
        \cline{1-2}
        \cline{7-7}
        Union14M-L~\cite{Union14M} & train & & & & & 4.1M \\
        \cline{1-2}
        \cline{7-7}
        OOV~\cite{OOV} & train & & & & & 4.4M \\
        \cline{1-2}
        \cline{7-7}
        LAION-OCR~\cite{Schuhmann2022LAION5BAO}\raisebox{-0.5ex}{\textsuperscript{*}} & train & & & & & 11.2M \\
        \cline{1-2}
        \cline{4-4}
        \cline{7-7}
        IAM~\cite{IAM} & trainval & & \multirow{3}{*}{handwritten} & & & 60K \\
        \cline{1-2}
        \cline{7-7}
        CVL~\cite{CVL} & train & & & & & 86K \\
        \cline{1-1}
        \cline{7-7}
        RIMES~\cite{RIMES} & trainval & & & & & 52K \\
        \cline{1-4}
        \cline{7-7}
        CROHME~\cite{CROHME} & trainval & \multirow{4}{*}{Cropped formula} & \multirow{4}{*}{none} & & & 11K \\
        \cline{1-2}
        \cline{7-7}
        HME100K~\cite{HME100K} & train & & & & & 75K \\
        \cline{1-2}
        \cline{7-7}
        LatexOCR\tablefootnote{Results are from: https://github.com/lukas-blecher/LaTeX-OCR
    } & trainval & & & & & 165K \\
        \cline{1-2}
        \cline{7-7}
        Synth-arxiv$\dagger$ & train & & & & & 10.2M \\
        \hline
        Total & - & - & - & \multicolumn{2}{|c|}{-} & 51.1M \\
        \hline
        \end{tabular}
    }
    \label{tab:comprehensive OCR datasets}
\end{table}

\nocite{IC13}
\nocite{IC15}
\nocite{ctw1500}
\nocite{totaltext}
\nocite{Long2022HierText}
\nocite{TextOCR}
\nocite{OpenVINO}
\nocite{Uber}
\nocite{COCO}
\nocite{2021cvpr_liu_abcnetv2}
\nocite{MLT17}
\nocite{Schuhmann2022LAION5BAO}
\nocite{Zhong2019PubLayNetLD}
\nocite{MJ1}
\nocite{syntext800k}
\nocite{Union14M}
\nocite{OOV}
\nocite{Schuhmann2022LAION5BAO}
\nocite{IAM}
\nocite{CVL}
\nocite{RIMES}
\nocite{CROHME}
\nocite{HME100K} 

\end{document}